\pdfoutput=1

\documentclass[11pt]{article}

\usepackage{ACL2023}
\usepackage{times}
\usepackage{latexsym}

\usepackage[T1]{fontenc}

\usepackage[utf8]{inputenc}

\usepackage{microtype}

\usepackage{subfigure}
\usepackage{graphicx}
\usepackage{amsmath}
\usepackage{amssymb}
\usepackage{booktabs}
\usepackage{multirow}
\usepackage{pifont}
\usepackage{tikz}
\usepackage{enumitem}

\definecolor{green}{RGB}{0, 153, 51}
\setitemize[1]{leftmargin=10pt,itemsep=0.5pt,partopsep=0pt,parsep=0.5pt,topsep=0pt}
\title{Medical Dialogue Generation via Dual Flow Modeling}

\author{Kaishuai Xu$^{1}$, Wenjun Hou$^{1,2\ast}$, Yi Cheng$^{1\ast}$, Jian Wang$^{1}$, Wenjie Li$^{1}$\\
$^1$Department of Computing, The Hong Kong Polytechnic University, Hong Kong \\
$^2$Research Institute of Trustworthy Autonomous Systems and \\Department of Computer Science and Engineering, \\
Southern University of Science and Technology, Shenzhen, China \\
\texttt{\{kaishuaii.xu, alyssa.cheng, jian-dylan.wang\}@connect.polyu.hk,} \\
\texttt{houwenjun060@gmail.com, cswjli@comp.polyu.edu.hk}
}

\begin{document}
\maketitle
\begingroup\def\thefootnote{$\ast$}\footnotetext{Equal Contributions.}\endgroup
\begin{abstract}
Medical dialogue systems (MDS) aim to provide patients with medical services, such as diagnosis and prescription. Since most patients cannot precisely describe their symptoms, dialogue understanding is challenging for MDS. 
Previous studies mainly addressed this by extracting the mentioned medical entities as critical dialogue history information. 
In this work, we argue that it is also essential to capture the transitions of the medical entities and the doctor's dialogue acts in each turn, as they help the understanding of how the dialogue flows and enhance the prediction of the entities and dialogue acts to be adopted in the following turn. 
Correspondingly, we propose a Dual Flow enhanced Medical (\textsc{DFMed}) dialogue generation framework. It extracts the medical entities and dialogue acts used in the dialogue history and models their transitions with an entity-centric graph flow and a sequential act flow, respectively. We employ two sequential models to encode them and devise an interweaving component to enhance their interactions. 
Experiments on two datasets demonstrate that our method exceeds baselines in both automatic and manual evaluations.

\end{abstract}

\section{Introduction}

Medical dialogue systems (MDS) have drawn considerable research attention with the increasing demand for telemedicine, especially after the outbreak of the COVID-19 pandemic \cite{meddialog,meddg, covid-19-meddialog,dialmed,ga-diag,remedi}, as they can provide much more people with in-time and affordable access to medical services such as health consultation, diagnosis, and prescription. 

\begin{figure}[th!]
	\centering
	\includegraphics[width=1\linewidth]{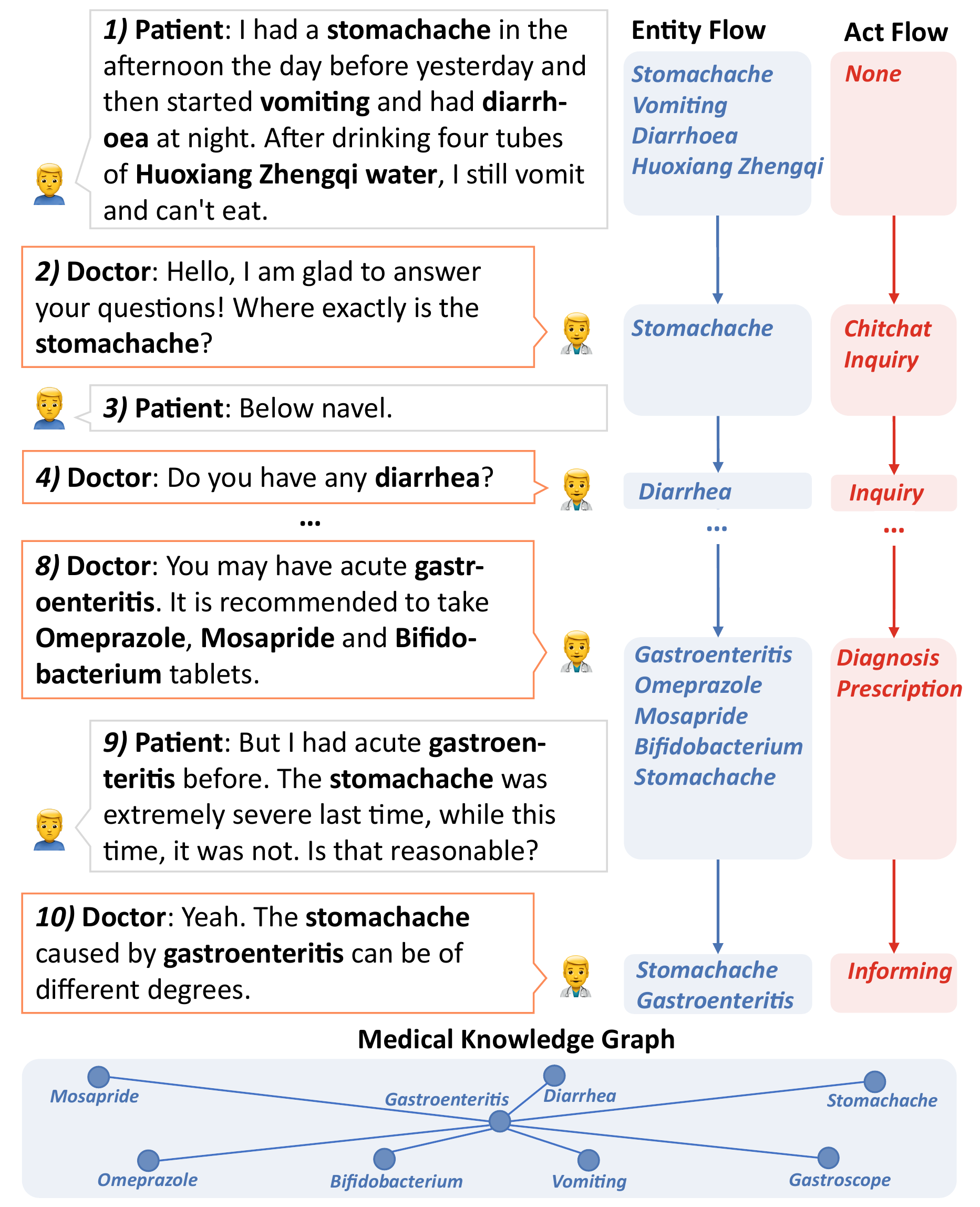}
	\caption{An example of a medical dialogue. \textbf{\textit{Diagnosis}} and \textbf{\textit{Prescription}} are short for \textit{Make a diagnosis} and \textit{Prescribe medications}, respectively.}
	\label{intro_example}
\end{figure}

For MDS, an efficient understanding of the dialogue history is challenging, as patients usually cannot describe their symptoms precisely and tend to convey lots of redundant information unnecessary for diagnosis \cite{meddg,fundamentals_of_clinical}. 
To extract the critical information in the lengthy dialogue, previous research focused on identifying the important medical entities mentioned in the context, such as diseases, medicine, and symptoms \cite{meddg,geml,hetero}. 

In our work, we argue that capturing the transitions of the medical entities and the doctor's dialogue acts in each turn (as depicted in Figure \ref{intro_example}) is also essential for the construction of MDS, which was largely overlooked by previous studies. 
In medical dialogues, the flows of medical entities and dialogue acts both follow particular transition patterns. For the medical entity flow, entities to appear in the following utterance are usually closely related to the ones mentioned recently. As in Figure \ref{intro_example}, the entities mentioned in adjacent dialogue rounds are logically related, being neighboring nodes in the medical knowledge. 
For the dialogue act flow, though variations are allowed to some extent, it usually needs to follow the medical consultation framework suggested in \newcite{skills_for_medical_com}. Modeling these two types of transitions would be helpful in dialogue understanding as they effectively capture how the dialogue history flows. 
Moreover, learning their transition patterns would also enhance the prediction of the dialogue acts and the medical entities to be adopted in the future turn.

Based on the above intuition, we propose a \textbf{D}ual \textbf{F}low enhanced \textbf{Med}ical (\textsc{DFMed}) dialogue generation framework. 
At each dialogue turn, it extracts the medical entities and the dialogue acts used in the dialogue history, and models their transitions with an entity-centric graph flow and a sequential act flow, respectively. 
Two sequential models are constructed to encode their transitions, with an interweaving component to enhance their interactions. 
The output representations are then used to predict the entities and the acts to be adopted in the following turn, which are employed to guide the response generation through gate control.  
Our main contributions are summarized as follows:
\begin{itemize}
    \item We propose a novel MDS framework, \textsc{DFMed}, which models the transitions of medical entities and dialogue acts via step-by-step interweaving. 
    \item We summarize the dialogue acts in the medical consultation scenario grounded on 
the medical documentation standards, SOAP notes \cite{soap}, including \emph{make a diagnosis}, \emph{prescribe medications}, etc. 
    \item Experimental results show the superiority of \textsc{DFMed} over the previous frameworks and demonstrate the effectiveness of introducing the medical entity and dialogue act flows. 
\end{itemize}

\section{Related Work}

Medical dialogue systems aim to provide medical services for patients. Early studies focus on automatic diagnosis in the form of a task-oriented dialogue system, and the purpose is to collect hidden symptoms in minimal turns and make a diagnosis at the end \cite{tod-diagnosis-hrl,symp-graph,diaformer,nose_run}. \newcite{tod-diagnosis} creates a dataset annotated with symptom phrases and constructs a reinforcement learning based MDS. \newcite{kg-routed-diagnosis} improves the topic transition in MDS by introducing a medical knowledge graph. With the release of large-scale medical dialogue datasets (e.g., MedDialog \cite{meddialog}, MedDG \cite{meddg}, and KaMed \cite{vrbot}), dialogue response generation attracts increasing attention. \newcite{meddg} frames medical dialogue generation as entity prediction and entity-aware response generation. Furthermore, \newcite{hetero} unifies the dialogue context understanding and entity reasoning through a heterogeneous graph. \newcite{vrbot} considers medical entities in patient and doctor utterances as states and actions and presents semi-supervised variation reasoning with a patient state tracker and a physician action network. The proposed model, VRBot, achieves comparable performance without entity supervision. \newcite{geml} analyses a low-resource challenge in medical dialogue generation and develops an entity-involved meta-learning framework to enhance diagnostic experience sharing between different diseases.

Although many studies have tried to improve medical dialogue generation by incorporating predicted medical entities, they simply treat entities in different turns as nodes in one entity graph with no entity transition modeling. Besides, few works focus on sequential entity-guided dialogue act prediction and sequential act-involved entity selection. Our framework exploits the transition and interaction of entity and act flows to strengthen dialogue understanding and guide response generation.

\section{Preliminary}

\paragraph{Problem Formulation.}
We define a medical dialogue as $U$$=$$\{(P_k, D_k)\}_{k=1}^T$, where $P$ and $D$ represent utterances from patients and doctors. Each utterance contains several medical entities $E$$=$$\{e_i\}$, and each doctor utterance is annotated with multiple dialogue acts $A$$=$$\{a_j\}$. Given the dialogue history $U_t$$=$$\{P_1, D_1, ..., P_t\}$, the system is supposed to generate the $t$-th doctor utterance $D_t$.

\paragraph{Dialogue Acts.}
We summarize several common dialogue acts implied in medical dialogues. One type is medical-related dialogue acts. We design acts that represent a function in a medical documentation standard, the SOAP note \cite{soap}, and occur throughout the dialogue. 
For example, \textit{State a required medical test} and \textit{Prescribe medications} in the ``Plan'' function of the SOAP note are included in our designed acts. The other type is general open-domain dialogue acts. We choose some acts introduced by \newcite{actflow} and further refine them, such as merging acts that behave as social obligation management into \textit{Chitchat}. Eventually, we obtain 7 dialogue acts for flow modeling, and later experiments demonstrate the effectiveness of guidance for response generation. Detailed description of each dialogue act can be found in Appendix \ref{act_detail}.

\section{Method}

\begin{figure}[t!]
	\centering
	\includegraphics[width=1\linewidth]{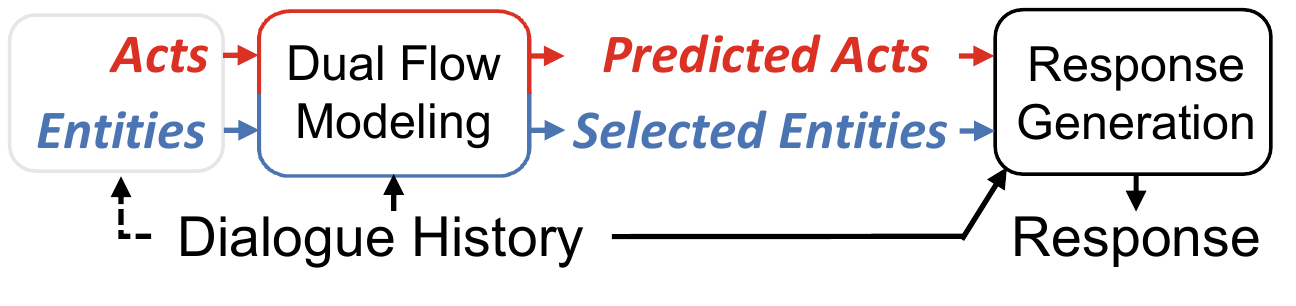}
	\caption{The overall framework of \textsc{DFMed}.}
	\label{frame}
\end{figure}

The proposed method learns two flows (i.e., a medical entity flow and a dialogue act flow) to model the propagation of medical entities and medical-related interactions and guide response generation with the corresponding hints. As shown in Figure \ref{frame}, the overall framework contains two modules. The Dual Flow Modeling module learns the entity and act transitions and infers probable entities and acts. The Response Generation module outputs a response under the guidance of the selected entities and predicted acts. 

\subsection{Dual Flow Modeling}

\begin{figure*}[th!]
	\centering
	\includegraphics[width=1\linewidth]{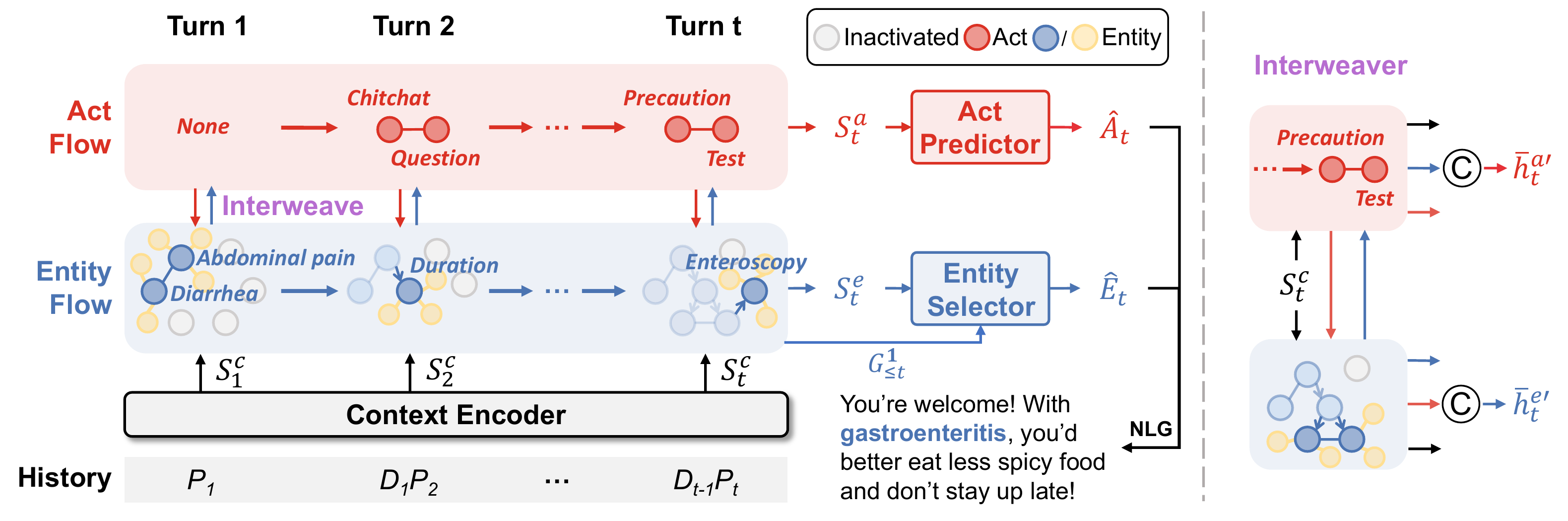}
	\caption{\textbf{Left}: The architecture of Dual Flow Modeling. It includes the dialogue act flow modeling in red and the medical entity flow modeling in blue. The one-hop sub-graph nodes of entities in dialogue are marked in yellow. \textbf{Right}: Detailed structure for flow interweaving. Each turn's graph embedding attends to previous act embeddings, and each turn's sequence embedding attends to previous entity embeddings.}
	\label{dualflow}
\end{figure*}

Figure \ref{dualflow} left displays the architecture of the Dual Flow Modeling module. To extract flow transitions, we encode the medical entities and the dialogue acts in a sequential way. Besides, at each turn, entity and act embeddings are mutually integrated through an interweaving component, named Interweaver. The context states $S^c$ produced by a context encoder also play a role in the integration. After encoding the whole dialogue history, we obtain the entity state $S^e_t$ and act state $S^a_t$. The entity states are adopted to select the relevant entities. The act states are sent to a multi-label act predictor to estimate the next acts. 

We apply BERT \cite{bert} as the context encoder to encode the dialogue history. All utterances are concatenated with a unique role token (i.e., ``[P]'' or ``[D]'') to separate the patient and doctor utterances. We compute the context states with different history ranges $U_k$$=$$\{P_1, D_1,..., P_k\} (k \in [1,t])$ for supporting the interweaving at each turn. The mean embedding of tokens in history $U_k$ is applied as the $k$-th turn context state $S^c_k \in \mathbb{R}^d$, where $d$ is the dimension of the state.

\subsubsection{Entity Flow}\label{entity_flow}

To learn the sequential transition of medical entities, we create the Entity Flow that encodes an entity graph transition and selects the most relevant entities. Specifically, the flow consists of an entity graph for each turn. Medical entities of each dialogue turn and entities from an external knowledge graph with a one-hop connection to the entities in dialogue are included in the graph of the $k$-th turn, denoted as $G_k$. Since we assume entities hop along the links on the graph, the ones in the sub-graph partially provide future transition hints. 

We define graphs that include entities until the $t$-th turn as $G_{\leq t}$. For all entities in graph $G_{\leq t}$, we use the same encoder as above to get entity embeddings. Instead of randomly initializing an embedding, BERT-based encoding keeps token-level semantics. 
Then, the average token embedding of each entity is employed as the raw embedding, denoted as $h^{e_0} \in \mathbb{R}^d$. Finally, Graph Attention Network (GAT) \cite{gat} is implemented to merge neighboring information for each entity:
\begin{align}
 \alpha^k_{ij} &= \frac{\exp \left(\sigma_1 \left(a^{\mathsf{T}} [W^k h_i^{e_0} \Vert W^k h_j^{e_0} ] \right) \right)}
 {\sum_{\mu \in \mathcal{N}_i}\exp \left(\sigma_1 \left(a^{\mathsf{T}} [W^k h_i^{e_0} \Vert W^k h_{\mu}^{e_0} ] \right) \right)}, \\
 h^e_i &= \left[ \sigma_2 \left(\sum_{j \in \mathcal{N}_i} \alpha^k_{ij} W^k h_j^{e_0} \right) \right]_{k=1}^K,
\end{align}
where $h^e_i \in \mathbb{R}^d$ is the updated entity embedding, $K$ is the number of heads, $a \in \mathbb{R}^{2d}$ is a trainable weight, $W^k \in \mathbb{R}^{d_h \times d}$ is the linear transformation matrix for the $k$-th head, $\sigma_1$ and $\sigma_2$ are the activation function, and $\mathcal{N}_i$ represents neighboring entities that connect to entity $i$ with one hop. The updated embedding is used to compute each turn's overall graph embedding via mean pooling:
\begin{align}
\bar{h}^e_k = \frac{1}{|G_k|} \sum_{i \in G_k} h^e_i, k\in[1, t],
\end{align}
where $\bar{h}^e_k \in \mathbb{R}^d$ is the entity graph embedding, $t$ represents the turn of the target response. Following \newcite{crs}, we employ a GRU to model the entity graph transition throughout the dialogue. The GRU takes all the previous graph embeddings $\{\bar{h}^e_1, \bar{h}^e_2, ..., \bar{h}^e_t\}$ as input and produces the entity state as follows:
\begin{align}
S^e_t = \text{GRU}_{\text{Entity}}(S^e_{t-1}, \bar{h}^e_t),
\end{align}
where $S^e_t \in \mathbb{R}^d$ is the final hidden state of the GRU. We denote $S^e_t$ as the entity state that implies clues for the entity transition in the previous context.

Then, we apply the entity state $S^e_t$ to calculate relevant scores for candidate entities. These entities are from the sub-graph of entities in a historical dialogue context. The score is defined as follows:
\begin{align}\label{score}
\text{Score} = \left\langle S^e_t, h^e_i \right\rangle, i \in G^{\textbf{1}}_{\leq t},
\end{align}
where $\langle, \rangle$ represents a similarity function, $h^e_i$ is the entity embedding, $G^{\textbf{1}}_{\leq t}$ is the one-hop sub-graphs for entities until the $t$-th turn. We select top-$k$ relevant entities $\hat{E}_t$ to guide response generation.

\subsubsection{Act Flow}\label{act_flow}

To learn the medical-related interactions of the doctor, we design the Act Flow that encodes act sequences and predicts the next acts. Specifically, the flow is composed of the act sequence of each turn. We first randomly initialize the trainable embedding of each dialogue act, denoted as $h^a\in \mathbb{R}^d$. The act sequence of the $k$-th turn can be defined as $A_k = \{h^a_1, h^a_2,...,h^a_{n^{k}_a}\}$, where $n^{k}_a$ represents the number of dialogue acts for each turn. Then, we compute the act sequence embedding $\bar{h}^a_k \in \mathbb{R}^d, k \in [1, t]$ through mean pooling.

Similar to the entity flow, we employ a GRU to model dialogue act transition. With all sequence embeddings as input, the final hidden state of the GRU is calculated and denoted as the act state:
\begin{align}
S^a_t = \text{GRU}_{\text{Act}}(S^a_{t-1}, \bar{h}^a_t),
\end{align}
where  $S^a_t \in \mathbb{R}^d$ is the act state of the $t$-th turn. Then, the multi-act probability of the $t$-th turn is computed based on the act state $S^a_t$ with a sigmoid and linear transformation layer:
\begin{align}\label{prob}
\text{Prob} = \text{sigmoid}(W_a S^a_t + b_a),
\end{align}
where $W_a \in \mathbb{R}^{n_a \times d}$ and $b_a \in \mathbb{R}^{n_a}$ are model parameters, and $n_a$ denotes the number of candidate dialogue acts. The predicted dialogue acts $\hat{A}_t$ are obtained through an appropriate threshold.

\subsubsection{Flow Interweaving}\label{interweave}
To achieve the integration of these two flows, we present an Interweaver to extend the entity graph embedding and act sequence embedding, as shown in Figure \ref{dualflow} right. This component incorporates the dialogue context into entity/act states and integrates entity/act sequential information from each other.

For the entity flow, we first fuse the historical context into the entity graph embedding. The context-aware graph embedding of each turn is computed via cross-attention:
\begin{align}
\alpha_{ki} &= \text{softmax}(\frac{{Q^c_k}^{\mathsf{T}} K^e_i}{\sqrt{d}}), \\
\bar{h}^{e^c}_k &= \sum_{i \in G_k} \alpha_{ki} V^e_i, k \in [1,t],
\end{align}
where $\bar{h}^{e^c}_k \in \mathbb{R}^d$ is the context-aware graph embedding at the $k$-th turn, $K^e_i$ and $V^e_i$ are linear projected vectors based on the entity embedding $h^e_i$, and $Q^c_k$ is based on the context state $S^c_k$. We denote this operation on $S^c_k$ and $[h^e_i]_{i \in G_k}$ as $\operatorname{CA}(S^c_k, [h^e_i]_{i \in G_k})$. Then, the act transition is fused into the graph embedding via $\operatorname{CA}$ as follows:
\begin{align}
\bar{h}^{e^a}_k = \operatorname{CA}(\bar{h}^{e}_k, [h^a_j]_{j \in A_{\leq k}}), k \in [1,t],
\end{align}
where $\bar{h}^{e^a}_k \in \mathbb{R}^d$ is the act-aware graph embedding, and $A_{\leq k}$ represents act sequences until the $k$-th turn. The final entity graph embedding is defined as the concatenation of three embeddings:
\begin{align}
\bar{h}^{e'}= [\bar{h}^e;\bar{h}^{e^c};\bar{h}^{e^a}],
\end{align}
where $\bar{h}_k^{e'} \in \mathbb{R}^{3d}$ is the extended graph embedding that incorporates historical dialogue context and act transition pattern.

For the act flow, we apply the $\operatorname{CA}$ operation to compute the context-aware and entity-aware act sequence embedding following the same way. The context-aware sequence embedding $\bar{h}^{a^c}_k$ is based on $S^c_k$ and $[h^a_j]_{j \in A_k}$, and the entity-aware sequence embedding $\bar{h}^{a^e}_k$ is based on $\bar{h}^{a}_k$ and $[h^e_i]_{i \in G_{\leq k}}$. These two embeddings incorporate the historical dialogue context and entity transition pattern. Then, the final sequence embedding $\bar{h}^{a'} \in \mathbb{R}^{3d}$ is concatenated as follows:
\begin{align}
\bar{h}^{a'}= [\bar{h}^a;\bar{h}^{a^c};\bar{h}^{a^e}].
\end{align}

We send the extended embeddings $\bar{h}^{e'}$ and $\bar{h}^{a'}$ instead of the pure embeddings to the above GRUs, allowing two flow models to acquire context information and lead each other.

\subsubsection{Training Objective}
The training of the dual flow modeling module includes two tasks, medical entity ranking and multi-act classification. The first one follows a contrastive learning \cite{simcse} way with a negative log likelihood loss:
\begin{align}
{\cal{L}}_e = - \sum_t^T \sum_{t^+} \log \frac{e^{\left\langle S^e_t, h_{t^+}^e \right\rangle}}{\sum_{i^- \in G_{\leq t}^{\textbf{1}}} e^{\left\langle S^e_t, h_{i^-}^e \right\rangle}},
\end{align}
where $h^e_{t^+}$ is the embedding of a target entity mentioned in the $t$-th response, $\langle , \rangle$ is the dot product operation that calculates relevant scores (see Eq. \ref{score}), and $G_{\leq t}^{\textbf{1}}$ denotes the one-hop sub-graphs for the dialogue history until the $t$-th turn. We randomly select several entities instead of the whole sub-graph as negative entities. The second one is defined as a multi-label classification with a binary cross-entropy loss:
\begin{align}
{\cal{L}}_a = \sum_t^T \sum^{n_a}_j \text{BCE}(\hat{y}_{t,j}^a, y_{t,j}^a),
\end{align}
where $\hat{y}_{t,j}^a$ is the probability of the dialogue act $j$ for the $t$-th response (see Eq. \ref{prob}), and $y_{t,j}^a$ is the ground-truth act label. The overall training objective of the dual flow modeling module can be calculated as:
\begin{align}
{\cal{L}}_{F} = \lambda_e {\cal{L}}_e + \lambda_a {\cal{L}}_a,
\end{align}
where $\lambda_e$ and $\lambda_a$ are weights for each task.

\subsection{Flow-guided Response Generation}

\begin{figure}[t]
	\centering
	\includegraphics[width=1\linewidth]{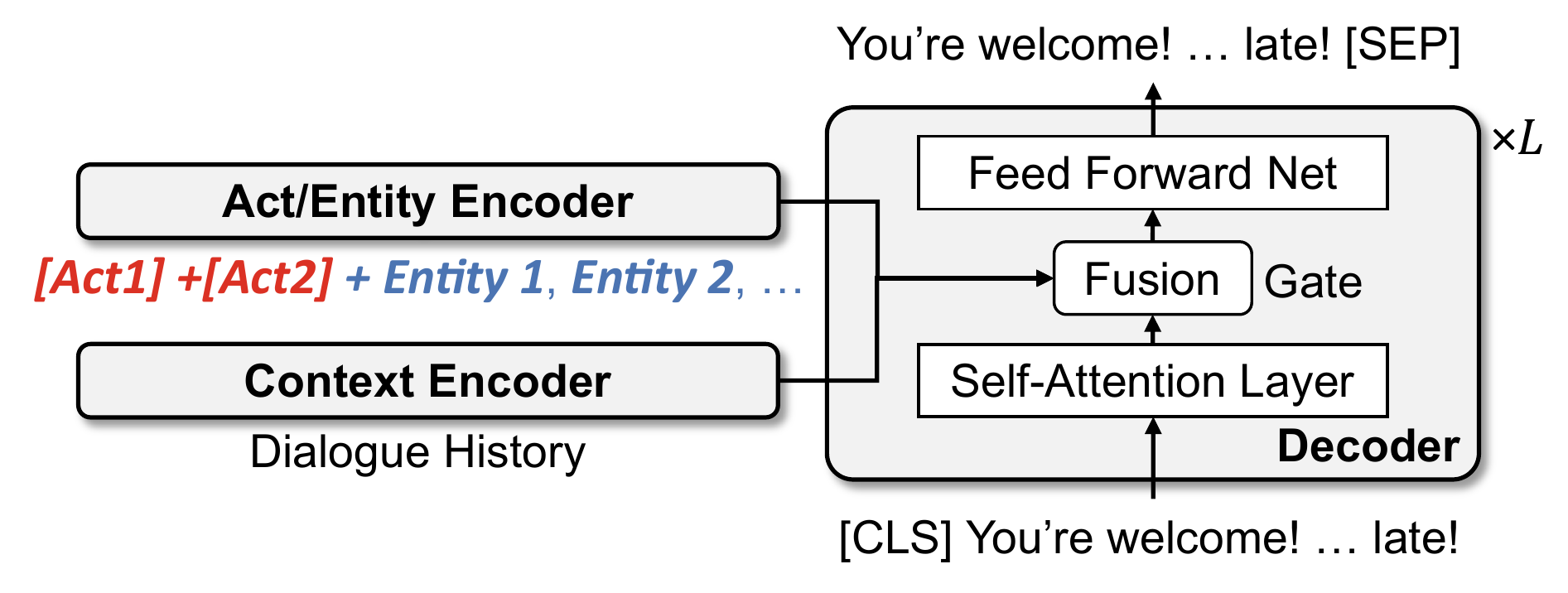}
	\caption{The detail of Response Generation.}
	\label{generator}
\end{figure}

After training the dual flow modeling module, we exploit the top-$k$ relevant entities $\hat{E}^t$ (see Sec. \ref{entity_flow}) and predicted dialogue acts $\hat{A}^t$ (see Sec. \ref{act_flow}) to guide response generation. We first encode the entity/act and dialogue history separately, allowing a relatively complete dialogue context. Then, these two types of information is merged into the decoder via a fusion component. 

\paragraph{Act-Entity Fusion}
As shown in Figure \ref{generator}, the dialogue acts and entities are concatenated into one token sequence. We assign a unique token to each dialogue act. Given the entity/act sequence and dialogue history sequence as input, the corresponding encoder produces final hidden states, $H^{ea}$ and $H^c$. We fuse two types of information through a gate mechanism in the decoder:
\begin{align}
h^{l, ea}_{w} &= \operatorname{CA}(h^l_{w}, H^{ea}), \\
h^{l,c}_{w} &= \operatorname{CA}(h^l_{w}, H^c), \\
g^{l}_{w} &= \text{sigmoid}(W^l h^{l,c}_{w}), \\
h^{l'}_{w} &= \text{FFN}(g^{l}_{w} h^{l, c}_{w} + (1 - g^{l}_{w}) h^{l, ea}_{w}),
\end{align}
where $h^{l'}_{w}$ is the final hidden state at the $w$-th token after fusion, $h^{l}_{w}$ denotes the output of the $l$-th self-attention layer, $\text{FFN}$ is the feed forward network, and $W^l$ is the trainable parameter. We adopt the final hidden state $h^{L'}_{w}$ of the decoder to compute the probability distribution of the next token:
\begin{align}
p(D_{t, w + 1}) = \text{softmax} (W_d h^{L'}_{w} + b_d),
\end{align}
where $W_d$ and $b_d$ are linear mapping parameters.

\paragraph{Training and Inference}

When training the response generation module, we use the selected top-$k$ entities from the dual flow modeling module and ground truth dialogue acts as encoder input. The training objective is defined as follows:
\begin{align}
{\cal{L}}_{G} = - \sum_t^T \sum_{w = 1}^{w}\log p(D_{t, w + 1}),
\end{align}
where $p(D_{t, w + 1})$ denotes the probability of the next token in the $t$-th response. Then in inference, we apply the top-$k$ entities and predicted dialogue acts to generate the next response.

\section{Experiments}

\subsection{Dataset}

Our experiments are conducted on two medical dialogue datasets, MedDG \cite{meddg} and KaMed \cite{vrbot}. 
The MedDG dataset contains 17K dialogues, focusing on 12 diseases in the gastroenterology department. The medical entities mentioned in the dialogues are annotated in MedDG. 
We split the dataset into 14862/1999/999 dialogues as the training/validation/test sets, following its original division. 
The KaMed dataset contains over 63K dialogues, covering diverse diseases in about 100 hospital departments. 
We filter out some dialogues with privacy concerns in KaMed (see Appendix \ref{process_kamed}) and obtain 29,159/1,532/1,539 dialogues as the training/validation/test sets. The final data accounts for around 51\% of the total.

\subsection{Baseline models}

We compare \textsc{DFMed} with five baseline models.  
(1) \textbf{Seq2Seq} \cite{seq2seq} is an RNN-based sequence to sequence model with an attention mechanism. (2) \textbf{HRED} \cite{hred} is a hierarchical RNN that models a dialogue as a token sequence and an utterance sequence. (3) \textbf{GPT-2} \cite{gpt-2} is a transformer decoder-based language model. (4) \textbf{BART} \cite{bart} is a transformer-based encoder-decoder model. (5) \textbf{VRBot} \cite{vrbot} is a medical dialogue generation model with patient entity tracking and doctor entity learning.

For the experiments on the MedDG dataset, we extend Seq2Seq, HRED, GPT-2, and BART with entity hints, following \newcite{meddg}. Specifically, the extracted medical entities are appended at the end of the input sequence. 
In the following, these models with entity modeling are displayed with the \textbf{-Entity} suffix (e.g., Seq2seq-Entity).

\subsection{Evaluation Metrics}

\paragraph{Automatic Evaluation.}

We evaluate modules in the proposed framework separately. For the dual flow modeling module, the top-20 recall rate (R@20) of target entities and the weighted F1 score (Weighted-F1) of different dialogue acts considering the act imbalance are adopted. For the response generation module, BLEU \cite{bleu} and ROUGE \cite{rouge} scores in different n-grams (i.e., B-1, B-2, B-4, R-1, and R-2) are adopted to assess the response quality. Besides, we also measure the precision, recall, and F1 of entities (i.e., E-P, E-R, and E-F1) in responses to demonstrate the reliability following \newcite{meddg}.

\paragraph{Human Evaluation.}

\begin{table*}[t!]
\centering
\resizebox{0.9\textwidth}{!}{
\begin{tabular}{llcccccccc} 
\toprule
\multicolumn{2}{c}{\textbf{Methods}}                                     & \textbf{B-1} & \textbf{B-2}              & \textbf{B-4}              & \textbf{R-1}              & \textbf{R-2}      & \textbf{E-P}              & \textbf{E-R}              & \textbf{E-F1}              \\ 
\midrule
\multicolumn{1}{c}{\multirow{5}{*}{w/o Pre-training}} & Seq2Seq          & 28.55                            & 22.85                     & 15.45                     & 25.61                     & 11.24             & 16.79                     & 10.44                     & 12.88                      \\
\multicolumn{1}{c}{}                                  & Seq2Seq-Entity   & 29.13                            & 23.22                     & 15.66                     & 25.79                     & 11.42             &\textbf{23.79}                     & 15.89                     & 19.06                      \\
\multicolumn{1}{c}{}                                  & HRED             & 31.61                            & 25.22                     & 17.05                     & 24.17                     & 9.79              & 15.56                     & 10.12                     & 12.26                      \\
\multicolumn{1}{c}{}                                  & HRED-Entity      & 32.84                            & 26.12                     & 17.63                     & 24.26                     & 9.76              & 21.75                     & 15.33                     & 17.98                      \\
\multicolumn{1}{c}{}                                  & VRBot            & 29.69                            & 23.90                     & 16.34                     & 24.69                     & 11.23             & 18.67                     & 9.72                      & 12.78                      \\ 
\midrule
\multirow{8}{*}{w/ Pre-trained LM}                    & GPT-2            & 35.27                            & 28.19                     & 19.16                     & 28.74                     & 13.61             & 18.29                     & 14.45                     & 16.14                      \\
                                                      & GPT-2-Entity     & 34.56                            & 27.56                     & 18.71                     & 28.78                     & 13.62             & 21.27                     & 17.10                     & 18.96                      \\
                                                      & BART             &34.94        & 27.99 & 19.06 & 29.03 & \textbf{14.40}             & 19.97 & 14.29 & 16.66  \\
                                                      & BART-Entity      & 34.14        & 27.19 & 18.42 & 28.52 & 13.67             & 23.49 & 16.90 & 19.66  \\ 
\cmidrule{2-10}
                                                      & \textsc{DFMed}            & \textbf{42.56}$^\dag$                            & 
  
  \textbf{33.34}$^\dag$               & 
  
  \textbf{22.53}$^\dag$               & 
  
  29.31               &

  14.21 & 
  
  22.48               & 
  
  \textbf{22.84}$^\dag$               & \textbf{22.66}$^\dag$                      \\
                                                      & w/o Act Flow     & 36.79                            & 29.18                     & 19.81                     & \textbf{29.45}                     & 14.26             & 22.73                     & 21.70                      & 22.20                      \\
                                                      & w/o Entity Flow  & 42.14                            & 32.83                     & 21.95                     & 29.26                     & 13.73             & 16.86                     & 22.06                     & 19.11                      \\
                                                      & w/o Interweaving & 42.35        & 33.02 & 22.19 & 29.02 & 14.11             & 22.14 & 20.62 & 21.34  \\
\bottomrule
\end{tabular}
}
\caption{\label{meddg_results}Automatic evaluation results on MedDG. † denotes statistically significant differences ($p$ < 0.05).}
\end{table*}

\begin{table}[t!]
\centering
\resizebox{0.47\textwidth}{!}{
\begin{tabular}{lccccc} 
\toprule
\textbf{Methods} & \textbf{B-1}              & \textbf{B-2} & \textbf{\textbf{B-4}}     & \textbf{\textbf{R-1}}     & \textbf{\textbf{R-2}}      \\ 
\midrule
Seq2Seq                              & 23.52                     & 18.56        & 12.13                     & 23.56                     & 8.67                       \\
HRED                                 & 26.75                     & 21.08        & 13.91                     & 22.93                     & 7.80                       \\
VRBot                                & 30.04                     & 23.76        & 16.36                     & 18.71                     & 7.28                       \\
GPT-2                                & 33.76                     & 26.58        & 17.82                     & 26.80                     & 10.56                      \\
BART                                 & 33.62 & 26.43        & 17.64 & 27.91 & 11.43  \\ 
\midrule
\textsc{DFMed}                                & \textbf{40.20}$^\dag$                     & \textbf{30.97}$^\dag$        & \textbf{20.76}$^\dag$                     & \textbf{28.28}$^\dag$                     & \textbf{11.54}$^\dag$                      \\
w/o Act Flow                         & 35.47                     & 28.11        & 18.78                     & 27.97                     & 11.45                      \\
w/o Entity Flow                      & 39.14                     & 29.92        & 19.73                     &27.17                     & 10.47                      \\
w/o Interweaving                     & 39.34 & 30.45        & 20.38 &28.03  & 11.39  \\
\bottomrule
\end{tabular}
}
\caption{\label{kamed_results}Automatic evaluation results on KaMed. † denotes statistically significant differences compared to all baselines ($p$ < 0.05).}
\end{table}

We chose 100 cases at random and invited three annotators to evaluate them manually. Results of the proposed framework are compared with different baseline models. We use three metrics to evaluate all of the generated responses based on past research \cite{meddg,vrbot}: \textit{sentence fluency} (FLU), \textit{knowledge accuracy} (KC), and \textit{entire quality} (EQ). On a 5-point Likert scale, from 1 (worst) to 5 (best), three annotators are asked to score these responses.

\subsection{Implementation Details}

For all baselines, we implement the open-source algorithm following \newcite{meddg} and \newcite{vrbot}.
We use the MedBERT\footnote{https://github.com/trueto/medbert} pretrained in the medical domain as the backbone of the dual flow modeling module. To extract the medical entities in the dialogue history, we refer to the medical knowledge graph CMeKG\footnote{http://cmekg.pcl.ac.cn/} and extract the text spans that match the string of nodes in the graph. Entities with a one-hop connection to the historical entities are target ones for entity flow modeling. Besides, following \newcite{remedi}, to identify the dialogue acts, we apply an open-source pseudo-labeling algorithm\footnote{https://github.com/yanguojun123/Medical-Dialogue} to automatically label each utterance. 
We train the dual flow modeling module with the AdamW \cite{adamw} optimizer. The learning rate and batch size are 4e-5 and 12 with 1000 warmup steps. The best loss weights $\lambda_e$ and $\lambda_a$ are 1 and 0.05 through grid searching. After training ten epochs, checkpoints with the highest average F1 for act prediction and the highest recall rate for top-20 entity selection on the validation set are selected. We select the threshold for each dialogue act, which achieves the best F1 score for the corresponding act on all validation samples. 
Then, we use Chinese pre-trained $\text{BART}_{base}$\footnote{https://huggingface.co/fnlp/bart-base-chinese} model with a six-layer encoder and a six-layer decoder for the response generation module. The entity/act encoder and context encoder share the same encoder. We adopt the AdamW optimizer and set the learning rate to 3e-5 with 2000 warmup steps. The model is trained for ten epochs with a batch size of 4. We implement all experiments on a single RTX 3090 GPU.

\begin{table}[t!]
\begin{center}
\resizebox{0.55\linewidth}{!}{
\begin{tabular}{@{}lccc@{}}
\toprule
\multicolumn{1}{c}{\textbf{Methods}} & \textbf{FLU} & \textbf{KC} & \textbf{EQ} \\ \midrule
BART                     &3.82           &1.86          &3.06          \\
BART-Entity              &3.85           &2.03          &3.35          \\
\textsc{DFMed}                 & \textbf{4.00}         & \textbf{2.14}        & \textbf{3.61}        \\ 
\midrule
Gold                 & 4.12         & 3.97        & 4.35        \\ 
\bottomrule
\end{tabular}
}
\end{center}
\caption{\label{human_eval} Human evaluation results on MedDG.}
\end{table}

\section{Results and Analysis}

\subsection{Automatic Evaluation}

\begin{table*}[t!]
\centering
\resizebox{0.77\linewidth}{!}{
\begin{tabular}{lcccccc} 
\toprule
\multirow{2}{*}{\textbf{Methods}} & \multicolumn{3}{c}{MedDG}                                           & \multicolumn{3}{c}{KaMed}                                           \\ 
\cline{2-7}
                                  & \textbf{Weighted-F1} & \textbf{R@20}        & \textbf{\textbf{B-4}} & \textbf{Weighted-F1} & \textbf{R@20}        & \textbf{B-4}          \\ 
\hline
\textsc{DFMed}                             & \textbf{62.83}       & \textbf{56.53}       & \textbf{22.53}        & \textbf{55.81}       & \textbf{52.23}       & \textbf{20.76}        \\
w/o Flow Modeling                 &62.09  &54.74  &22.11   &55.16  &51.31  &20.21   \\
w/o Interweaving                  & 62.13                & 54.98                & 22.19                 & 55.17                & 51.44                & 20.38                 \\
w/o Entity attends to Act                 & 62.37                & 55.75                & 22.34                 & 55.43                & 51.84                & 20.62                 \\
w/o Act attends to Entity                 & 62.43                & 55.91                & 22.40                 & 55.62                & 52.10                & 20.67                 \\
\bottomrule
\end{tabular}
}
\caption{\label{ablation} Results of Dual Flow Modeling with ablation. The best results are in \textbf{bold}.}
\end{table*}

The overall comparison of \textsc{DFMed} and other baseline models on the MedDG dataset is illustrated in Table \ref{meddg_results}, and the KaMed dataset is in Table \ref{kamed_results}. The observations from the comparison are as follows:

(1) Our proposed framework \textsc{DFMed} outperforms these baseline models in most metrics. Specifically, on the MedDG dataset, it is 8.42\%, 6.15\%, 4.11\%, and 0.79\% higher than the best baseline model, BART-Entity, on B-1, B-2, B-4, and R-1. On the KaMed dataset, \textsc{DFMed} outperforms BART by 6.58\%, 4.54\%, 3.12\%, and 0.37\% on B-1, B-2, B-4, and R-1, indicating better similarity to the content of ground truth responses. Besides, on the MedDG dataset, \textsc{DFMed} exceeds BART-Entity by 3.00\% on E-F1, meaning that it can generate responses with more accurate entity mentions. The above increases are because \textsc{DFMed} learns the transition of a medical entity flow and a dialogue act flow and predicts the entities and acts to guide response generation. Moreover, interweaving two flows enhances the prediction of entities and acts.

(2) \textsc{DFMed} effectively fuses the guidance from medical entities and dialogue acts. Compared to GPT-2 and BART, which have a performance drop on BLEU with the incorporation of medical entities (i.e., \textbf{-Entity}), \textsc{DFMed} shows increases in all these metrics. The main reason is that the force incorporation of entities may reduce response fluency. In \textsc{DFMed}, the dialogue act guidance influences attention to entities in the encoder and implicitly prevents the enforcement. The gate mechanism in the decoder controls the proportion of information from entities and acts. Besides, dialogue acts also provide essential content for responses without entity hints. This improvement is significant as about half of the responses in datasets do not match an entity in CMeKG via automatic string matching.

\subsection{Human Evaluation}

 We select methods with high accuracy on the MedDG dataset to conduct a human evaluation, as shown in Table \ref{human_eval}. Our framework displays an overall better response quality. Especially on the EQ, \textsc{DFMed} performs significantly better than baselines due to the incorporation of dialogue acts. The Cohen's Kappa coefficient equals 0.53 and indicates a moderate agreement \cite{cohen_kappa}.

\subsection{Analysis of Dual Flow Modeling}

To further explore the effectiveness of our method, we investigate the following variants of \textsc{DFMed}: (1) \textbf{w/o Flow Modeling}, where we use a context state produced by mean pooling of all hidden states of dialogue context tokens to rank entities and predict dialogue acts. (2) \textbf{w/o Interweaving}, which removes the interweaving between entities and acts. (3) \textbf{w/o Entity attends to Act}, which removes the interweaving from act sequences. (4) \textbf{w/o Act attends to Entity}, which removes the interweaving from sequential entity graphs.

Table \ref{ablation} shows the ablation study results. We observe drops in all metrics with the ablation variants, indicating the effectiveness of our proposed module. Specifically, the result of \textit{w/o Flow Modeling} significantly drops on the entity recall rate and slightly drops on the overall act prediction F1 compared to the full model. It demonstrates that flow modeling can be necessary for learning the transition of entities and acts in dialogues. Besides, comparison among variants of the interweaver illustrates that incorporating sequential entity graphs and act sequences assists in the transition. Notably, the act sequence is more conducive to the transition of the entity and act flows. It may be because the entity graph introduce noise entities from patient utterances. These entities are more variable and can be deviated from the main content.

\subsection{Case Study}

\begin{table}[!th]
\centering
\small
\begin{tabular}{p{0.94\linewidth}}
\toprule
\textbf{Patient}: Hello, doctor! I feel a little \underline{stomachache} and \underline{suffocated}. The stomachache is intermittent, sometimes it hurts, sometimes it doesn't. When I am not active, the pain is not obvious, but when I move, bloating makes me sick. I still have \underline{diarrhea}. :( \\
\textbf{Doctor}: Hi, is it \underline{colic}? Does stomachache have anything to do with eating?                      \\
\textbf{Patient}: No! It hurts before and after eating.     \\
...                                                                         \\
\textbf{Patient}: I've never had a \underline{gastroscopy}.                                                                 \\ 
\midrule
\textbf{Gold Response}: You may have \underline{gastritis} or \underline{stomach} \underline{cramps} and need to do a \underline{gastroscopy} or a barium meal first! If diagnosed with the above diseases, you can take \underline{omeprazole}, \underline{Kangfuxin} liquid, and \underline{belladonna} tablets following specific instructions.         \\ 
\midrule
\textbf{BART}: I suggest you try some \underline{omeprazole} and \underline{hydrotalcite} chewable tablets.                                                  \\
\textbf{BART-Entity}: It is recommended to have a \underline{gastroscopy} to see if there is any history of \underline{gastritis} or \underline{gastric} \underline{ulcer}.                                                               \\
\midrule
\textbf{Selected Entities}: gastritis, gastroscopy, omeprazole,...; \\
\textbf{Predicted Acts}: \textcolor{red}{[}diagnosis\textcolor{red}{]}, \textcolor{blue}{[}prescription\textcolor{blue}{]}, \textcolor{green}{[}test\textcolor{green}{]}. \\
\textbf{Ours (\textsc{DFMed})}: \textcolor{red}{[}You may have \underline{gastritis} or gastroesophageal reflux.\textcolor{red}{]} \textcolor{blue}{[}It is recommended to take \underline{omeprazole}, \underline{mosapride}, \underline{hydrotalcite}.\textcolor{blue}{]} \textcolor{green}{[}If the symptoms are not relieved, it is recommended to do a \underline{gastroscopy}.\textcolor{green}{]}                      \\
\bottomrule
\end{tabular}
\caption{\label{case_study} Case study. The responses are generated by different models, where key entities are underlined.}
\end{table}

A case generated by the above methods is illustrated in Table \ref{case_study}. Compared to baseline models, \textsc{DFMed} can produce responses more consistent with the dialogue context reflected by medical entities and dialogue acts. We can observe that the dual flow modeling module correctly predicts all acts and selects several medical entities, although two medications (e.g., ``mosapride'') are different from ground truth mentions. Then, the beneficial guidance from medical entities and acts is employed to generate the next response. The response generation module fuses the entities and acts and produces the response containing these two hints.

\section{Conclusion}

In this paper, we propose a dual flow enhanced medical dialogue generation framework, \textsc{DFMed}, that models a medical entity flow and a dialogue act flow to improve the relevant entity selection and dialogue act prediction. Besides, we design an interweaver to strengthen interactions of two flows. The selected entities and predicted acts are applied to guide response generation. Experiments validate the effectiveness of our \textsc{DFMed} on two datasets.

\section*{Limitations}

Although our proposed framework beats several baseline methods for medical dialogue generation, there is still room for progress. We exploit an entity flow and a dialogue act flow to improve dialogue understanding and guide response generation. However, our summarized dialogue acts are limited in the types and granularity of functions they denote. We can manually annotate more medical-related dialogue acts in our future research following the SOAP notes. Besides, more medical knowledge with different formats, such as medical articles and medical examination reports, can be incorporated. Finally, it is crucial to recognize the potential risks associated with system utilization and the possibility of patient privacy leakage. A collaborative approach involving both dialogue systems and medical professionals should be considered. This will ensure that responses are endorsed by physicians and stringently overseen by reliable authorities.

\section*{Ethics Statement}

Our proposed system aims to provide medical services, such as diagnosis and prescription, for patients who suffer from certain diseases. All datasets have been anonymized when released in dataset papers. However, since we train the model with limited and incomplete samples in two datasets, the generated responses may involve misleading information about diagnosis, treatment, and precautions. We recommend that users adopt the system as an auxiliary tool and go to the hospital for help if necessary. Besides, when interacting with the system, there is a risk of sensitive information leak (e.g., gender as reported by users). It can be addressed by adopting anonymous technology in the future. Thus, we strongly advise users to consider the ethical implications of the generated responses carefully. Furthermore, the scientific artifacts that we used are freely available for research, including NLTK, ROUGE, Transformers, and other GitHub codes. And this paper's use of these artifacts is consistent with their intended use.

\section*{Acknowledgment}
This work was supported by the Research Grants Council of Hong Kong (15207920, 15207821, 15207122) and National Natural Science Foundation of China (62076212). 

\bibliography{custom}
\bibliographystyle{acl_natbib}

\appendix

\section{Appendix}
\subsection{Details of Packages}

We use the NLTK package in version 3.4.1 for calculating BLEU scores, the Pyrouge package in version 0.1.3 for calculating ROUGE scores, and the Transformers package in version 4.21.3.

\subsection{Details of Dialogue Acts}\label{act_detail}

\begin{table}[th!]
\centering
\resizebox{0.95\linewidth}{!}{
\begin{tabular}{lrr} 
\toprule
\textbf{Dialogue Acts}        & MedDG & KaMed  \\ 
\hline
Inquire                       &25.49\%    &20.95\%        \\
Make a diagnosis              &6.72\%     &8.64\%        \\
Prescribe medications         &10.12\%    &13.74\%        \\
State a required medical test &4.25\%     &8.29\%        \\
Provide daily precautions     &7.51\%     &5.29\%        \\
Inform                        &29.91\%    &30.04\%        \\
Chitchat                      &15.98\%    &13.04\%        \\
\bottomrule
\end{tabular}
}
\caption{\label{act_proportion} The proportion of each dialogue act in the MedDG and KaMed datasets.}
\end{table}

In this section, we will describe the detail of our summarized dialogue acts. The seven dialogue acts can be divided into two types: (i) medical-related functions and (ii) general open-domain dialogue acts. The specific meaning of each act is interpreted as follows:

\paragraph{Medical-related functions.} (1) \textbf{Inquire}. The doctor asks questions about the history of the present illness (e.g., the location and duration of one symptom), previous surgery and medical conditions, current medications, allergies, etc. It corresponds to the ``Subjective'' function of the SOAP note \cite{soap}. 
(2) \textbf{Make a diagnosis}. The doctor makes a differential diagnosis based on historical dialogue context. It corresponds to the ``Assessment'' function of the SOAP note. (3) \textbf{Prescribe medications}. The doctor provide medication names and instructions. (4) \textbf{State a required medical test}. The doctor explains which tests are required and why each one was chosen to resolve diagnostic ambiguities; besides, what the next step would be if the results are positive or negative. (5) \textbf{Provide daily precautions}. The doctor explains the things that need to be paid attention to every day. Acts (3), (4), and (5) correspond to the ``Plan'' function of the SOAP note.

\paragraph{General open-domain dialogue acts.} (6) \textbf{Inform}. The doctor tells the patient some information that he assumes to be correct. (7) \textbf{Chitchat}. The doctor expresses welcome, goodbye, apology and thanks to the patient.

The proportion of each act in the MedDG and KaMed datasets is shown in the Table \ref{act_proportion}. Examples for each dialogue act are listed as follows:

\begin{enumerate}
    \item \textbf{Inquire}: ``Hello, do you usually have diarrhea?'', ``Have you taken any medicine before?'';
    \item \textbf{Make a diagnosis}: ``You may have gastroenteritis'', ``You may have allergic rhinitis, which is easy to get sick this season.'';
    \item \textbf{Prescribe medications}: ``Please take Amoxicillin capsule 1.0g 2 times a day and Clarithromycin tablet 0.5g 2 times a day. Both are taken after meals.'';
    \item \textbf{State a required medical test}: ``If you often feel sick to your stomach, you can do a gastroscopy.'', ``If your condition does not improve, I suggest you do a gastroscopy and Helicobacter pylori detection.'';
    \item \textbf{Provide daily precautions}: ``Please drink plenty of water, eat more fruits and vegetables. And try to have a morning poop.'';
    \item \textbf{Inform}: ``Migraine is a primary headache whose etiology and pathogenesis are not fully understood.'', ``It can be effective in five days, and individual differences are relatively large.'';
    \item \textbf{Chitchat}: ``You're welcome, and I wish you a speedy recovery!'', ``Thank you so much'', ``Hello!'';
\end{enumerate}

\subsection{Extra Process of the KaMed dataset}\label{process_kamed}

\begin{table}[!th]
\centering
\small
\begin{tabular}{p{0.94\linewidth}}
\toprule

\textbf{Patient}: Can I apply a facial mask if I have pimples on my face? \\
\textbf{Doctor}: Hello, do you have a picture to show? How long has it been? Are there any discomforts? \\
\textbf{Patient}: \underline{The image is not available for privacy concerns.} \\
\textbf{Doctor}: Based on your description, it seems like acne, also known as pimples or blackheads. \\
... \\

\bottomrule
\end{tabular}
\caption{\label{filter_sample} An example of the filtered dialogues.}
\end{table}

We filter out dialogues that include texts such as ``The image is not available for privacy concerns'' and ``The voice is not available for privacy concerns'' since these dialogues are incomplete and hard to understand. We have obtained approximately 51\% samples for our experiments. The reason is as follows.

The KaMed dataset is collected from an online medical consultation platform. The raw dialogues contain multi-modal information such as pictures (e.g., medical examination reports and photos of body parts) and voice messages. These messages are crucial for dialogue development since the doctor will respond to the picture or voice. For example, they will discuss the result of an examination report, or patients will directly use voice messages instead of texts to express their condition. However, this information is replaced by meaningless text such as "The image is not available" for privacy concerns when collecting the dataset. Thus, the dialogue context is incomplete and difficult to understand. We argue that filtering dialogues that contain these texts can help us build a more robust model for dialogue understanding.

\end{document}